# Belief Propagation for Structured Decision Making

**Qiang Liu**
Department of Computer Science
University of California, Irvine
Irvine, CA, 92697
qliu1@ics.uci.edu

**Alexander Ihler**
Department of Computer Science
University of California, Irvine
Irvine, CA, 92697
ihler@ics.uci.edu

## Abstract

Variational inference algorithms such as belief propagation have had tremendous impact on our ability to learn and use graphical models, and give many insights for developing or understanding exact and approximate inference. However, variational approaches have not been widely adoped for *decision making* in graphical models, often formulated through influence diagrams and including both centralized and decentralized (or multi-agent) decisions. In this work, we present a general variational framework for solving structured cooperative decision-making problems, use it to propose several belief propagation-like algorithms, and analyze them both theoretically and empirically.

## 1 Introduction

Graphical modeling approaches, including Bayesian networks and Markov random fields, have been widely adopted for problems with complicated dependency structures and uncertainties. The problems of *learning*, i.e., estimating a model from data, and *inference*, e.g., calculating marginal probabilities or maximum *a posteriori* (MAP) estimates, have attracted wide attention and are well explored. Variational inference approaches have been widely adopted as a principled way to develop and understand many exact and approximate algorithms. On the other hand, the problem of *decision making* in graphical models, sometimes formulated via influence diagrams or decision networks and including both sequential centralized decisions and decentralized or multi-agent decisions, is surprisingly less explored in the approximate inference community.

*Influence diagrams* (ID), or *decision networks*, [Howard and Matheson, 1985, 2005] are a graphical model representation of structured decision problems under uncertainty; they can be treated as an extension of Bayesian networks, augmented with decision nodes and utility functions. Traditionally, IDs are used to model centralized, sequential decision processes under "perfect recall", which assumes that the decision steps are ordered in time and that all information is remembered across time; limited memory influence diagrams (LIMIDs) [Zhang et al., 1994, Lauritzen and Nilsson, 2001] relax the perfect recall assumption, creating a natural framework for representing decentralized and information-limited decision problems, such as team decision making and multi-agent systems. Despite the close connection and similarity to Bayes nets, IDs have less visibility in the graphical model and automated re-seasoning community, both in terms of modeling and algorithm development; see Pearl [2005] for an interesting historical perspective.

Solving an ID refers to finding decision rules that maximize the expected utility function (MEU); this task is significantly more difficult than standard inference on a Bayes net. For IDs with perfect recall, MEU can be restated as a dynamic program, and solved with cost exponential in a constrained tree-width of the graph that is subject to the temporal ordering of the decision nodes. The constrained tree-width can be much higher than the tree-width associated with typical inference, making MEU significantly more complex. For LIMIDs, non-convexity issues also arise, since the limited shared information and simultaneous decisions may create locally optimal policies. The most popular algorithm for LIMIDs is based on policy-by-policy improvement [Lauritzen and Nilsson, 2001], and provides only a "person-by-person" notion of optimality. Surprisingly, the variational ideas that revolutionized inference in Bayes nets have not been adopted for influence diagrams. Although there exists work on transforming MEU problems into sequences of standard marginalization problems [e.g., Zhang, 1998], on which variational methods apply, these methods do not yield general frameworks, and usually only work for IDs with perfect recall. A full variational frame-work would provide general procedures for developing efficient approximations such as loopy belief propagation (BP), that are crucial for large scale problems, or providing new theoretical analysis.

In this work, we propose a general variational framework for solving influence diagrams, both with and without perfect recall. Our results on centralized decision making include traditional inference in graphical models as special cases. We propose a spectrum of exact and approximate algorithms for MEU problems based on the variational framework. We give several optimality guarantees, showing that under certain conditions, our BP algorithm can find the globally optimal solution for ID with perfect recall and solve LIMIDs in a stronger locally optimal sense than coordinate-wise optimality. We show that a temperature parameter can also be introduced to smooth between MEU tasks and standard (easier) marginalization problems, and can provide good solutions by annealing the temperature or using iterative proximal updates.

This paper is organized as follows. Section 2 sets up background on graphical models, variational methods and influence diagrams. We present our variational framework of MEU in Section 3, and use it to develop several BP algorithms in Section 4. We present numerical experiments in Section 5. Finally, we discuss additional related work in Section 6 and concluding remarks in Section 7. Proofs and additional information can be found in the appendix.

## 2 Background

### 2.1 Graphical Models

Let $\boldsymbol{x} = \{x_1, x_2, \cdots, x_n\}$ be a random vector in $\mathbb{X} = \mathbb{X}_1 \times \cdots \times \mathbb{X}_n$. Consider a factorized probability on $\boldsymbol{x}$,

$$p(\boldsymbol{x}) = \frac{1}{Z} \prod_{\alpha \in \mathcal{I}} \psi_\alpha(x_\alpha) \ = \ \frac{1}{Z} \exp\Big[\sum_{\alpha \in \mathcal{I}} \theta_\alpha(x_\alpha)\Big],$$

where $\mathcal{I}$ is a set of variable subsets, and $\psi_\alpha \colon \mathbb{X}_\alpha \to \mathbb{R}^+$ are positive factors; the $\theta_\alpha(x_\alpha) = \log \psi_\alpha(x_\alpha)$ are the natural parameters of the exponential family representation; and $Z = \sum_{\boldsymbol{x}} \prod_{\alpha \in \mathcal{I}} \psi_\alpha$ is the normalization constant or partition function with $\Phi(\boldsymbol{\theta}) = \log Z$ the log-partition function. Let $\boldsymbol{\theta} = \{\theta_\alpha | \alpha \in \mathcal{I}\}$ and $\theta(\boldsymbol{x}) = \sum_\alpha \theta_\alpha(x_\alpha)$. There are several ways to represent a factorized distribution using graphs (i.e., *graphical models*), including Markov random fields, Bayesian networks, factors graphs and others.

Given a graphical model, *inference* refers to the procedure of answering probabilistic queries. Important inference tasks include marginalization, maximum *a posteriori* (MAP, sometimes called maximum probability of evidence or MPE), and marginal MAP (sometimes simply MAP). All these are NP-hard in general. Marginalization calculates the marginal probabilities of one or a few variables, or equivalently the normalization constant $Z$, while MAP/MPE finds the mode of the distribution. More generally, marginal MAP seeks the mode of a marginal probability,

$$\text{Marginal MAP:} \quad \boldsymbol{x}^* = \arg\max_{x_A} \sum_{x_B} \prod_\alpha \psi_\alpha(x_\alpha),$$

where $A, B$ are disjoint sets with $A \cup B = V$; it reduces to marginalization if $A = \emptyset$ and to MAP if $B = \emptyset$.

**Marginal Polytope.** A marginal polytope $\mathbb{M}$ is a set of local marginals $\boldsymbol{\tau} = \{\tau_\alpha(x_\alpha) \colon \alpha \in \mathcal{I}\}$ that are extensible to a global distribution over $\boldsymbol{x}$, that is, $\mathbb{M} = \{\boldsymbol{\tau} | \ \exists$ a distribution $p(\boldsymbol{x})$, s.t. $\sum_{x_{V \setminus \alpha}} p(\boldsymbol{x}) = \tau_\alpha(x_\alpha)\ \}$. Call $\mathcal{P}[\boldsymbol{\tau}]$ the set of global distributions consistent with $\boldsymbol{\tau} \in \mathbb{M}$; there exists a unique distribution in $\mathcal{P}[\boldsymbol{\tau}]$ that has maximum entropy and follows the exponential family form for some $\boldsymbol{\theta}$. We abuse notation to denote this unique global distribution $\tau(\boldsymbol{x})$.

A basic result for variational methods is that $\Phi(\boldsymbol{\theta})$ is convex and can be rewritten into a dual form,

$$\Phi(\boldsymbol{\theta}) = \max_{\boldsymbol{\tau} \in \mathbb{M}} \{\langle \boldsymbol{\theta}, \boldsymbol{\tau}\rangle + H(\boldsymbol{x}; \boldsymbol{\tau})\}, \qquad (1)$$

where $\langle \boldsymbol{\theta}, \boldsymbol{\tau}\rangle = \sum_{\boldsymbol{x}} \sum_\alpha \theta_\alpha(x_\alpha) \tau_\alpha(x_\alpha)$ is the point-wise inner product, and $H(\boldsymbol{x}; \boldsymbol{\tau}) = -\sum_{\boldsymbol{x}} \tau(\boldsymbol{x}) \log \tau(\boldsymbol{x})$ is the entropy of distribution $\tau(\boldsymbol{x})$; the maximum of (1) is obtained when $\boldsymbol{\tau}$ equals the marginals of the original distribution with parameter $\boldsymbol{\theta}$. See Wainwright and Jordan [2008].

Similar dual forms hold for MAP and marginal MAP. Letting $\Phi_{A,B}(\boldsymbol{\theta}) = \log \max_{x_A} \sum_{x_B} \exp(\theta(\boldsymbol{x}))$, we have [Liu and Ihler, 2011]

$$\Phi_{A,B}(\boldsymbol{\theta}) = \max_{\boldsymbol{\tau} \in \mathbb{M}} \{\langle \boldsymbol{\theta}, \boldsymbol{\tau}\rangle + H(x_B | x_A\, ; \boldsymbol{\tau})\}, \qquad (2)$$

where $H(x_B | x_A\, ; \boldsymbol{\tau}) = -\sum_{\boldsymbol{x}} \tau(\boldsymbol{x}) \log \tau(x_B | x_A)$ is the conditional entropy; its appearance corresponds to the sum operators.

The dual forms in (1) and (2) are no easier to compute than the original inference. However, one can approximate the marginal polytope $\mathbb{M}$ and the entropy in various ways, yielding a body of approximate inference algorithms, such as loopy belief propagation (BP) and its generalizations [Yedidia et al., 2005, Wainwright et al., 2005], linear programming solvers [e.g., Wainwright et al., 2003b], and recently hybrid message passing algorithms [Liu and Ihler, 2011, Jiang et al., 2011].

**Junction Graph BP.** Junction graphs provide a procedural framework to approximate the dual (1). A cluster graph is a triple $(\mathcal{G}, \mathcal{C}, \mathcal{S})$, where $\mathcal{G} = (\mathcal{V}, \mathcal{E})$ is an undirected graph, with each node $k \in \mathcal{V}$ associated with a subset of variables $c_k \in \mathcal{C}$ (clusters), and each edge $(kl) \in \mathcal{E}$ a subset $s_{kl} \in \mathcal{S}$ (separator) satisfying $s_{kl} \subseteq c_k \cap c_l$. We assume that $\mathcal{C}$ subsumes the index set $\mathcal{I}$, that is, for any $\alpha \in \mathcal{I}$, there exists a $c_k \in \mathcal{C}$, denoted $c[\alpha]$, such that $\alpha \subseteq c_k$. In this case, we can reparameterize $\boldsymbol{\theta} = \{\theta_\alpha | \alpha \in \mathcal{I}\}$ into $\boldsymbol{\theta} = \{\theta_{c_k} | k \in \mathcal{V}\}$ by taking $\theta_{c_k} = \sum_{\alpha:\ c[\alpha]=c_k} \theta_\alpha$, without changing the distribution. A cluster graph is called a *junction graph* if it satisfies the *running intersection property* – for each $i \in V$, the induced sub-graph consisting of the clusters and separators that include $i$ is a connected tree. A junction graph is a junction tree if $\mathcal{G}$ is tree.

To approximate the dual (1), we can replace $\mathbb{M}$ with a locally consistent polytope $\mathbb{L}$: the set of local marginals $\boldsymbol{\tau} = \{\tau_{c_k}, \tau_{s_{kl}}: k \in \mathcal{V}, (kl) \in \mathcal{E}\}$ satisfying $\sum_{x_{c_k \backslash s_{kl}}} \tau_{c_k}(x_{c_k}) = \tau(x_{s_{kl}})$. Clearly, $\mathbb{M} \subseteq \mathbb{L}$. We then approximate (1) by

$$\max_{\boldsymbol{\tau} \in \mathbb{L}} \{\langle \boldsymbol{\theta}, \boldsymbol{\tau} \rangle + \sum_{k \in \mathcal{V}} H(x_{c_k}; \boldsymbol{\tau}_{c_k}) - \sum_{(kl) \in \mathcal{E}} H(x_{s_{kl}}; \boldsymbol{\tau}_{s_{kl}})\},$$

where the joint entropy is approximated by a linear combination of the entropies of local marginals. The approximate objective can be solved using Lagrange multipliers [Yedidia et al., 2005], leading to a sum-product message passing algorithm that iteratively sends messages between neighboring clusters via

$$m_{k \to l}(x_{c_l}) \propto \sum_{x_{c_k \backslash s_{kl}}} \psi_{c_k}(x_{c_k}) m_{\sim k \backslash l}(x_{c_{\mathcal{N}(k)}}), \quad (3)$$

where $\psi_{c_k} = \exp(\theta_{c_k})$, and $m_{\sim k \backslash l}$ is the product of messages into $k$ from its neighbors $\mathcal{N}(k)$ except $l$. At convergence, the (locally) optimal marginals are

$$\tau_{c_k} \propto \psi_{c_k} m_{\sim k} \quad \text{and} \quad \tau_{s_{kl}} \propto m_{k \to l} m_{l \to k},$$

where $m_{\sim k}$ is the product of messages into $k$. Max-product and hybrid methods can be derived analogously for MAP and marginal MAP problems.

### 2.2 Influence Diagrams

*Influence diagrams* (IDs) or *decision networks* are extensions of Bayesian networks to represent structured decision problems under uncertainty. Formally, an influence diagram is defined on a directed acyclic graph $G = (V, E)$, where the nodes $V$ are divided into two subsets, $V = R \cup D$, where $R$ and $D$ represent respectively the set of chance nodes and decision nodes. Each chance node $i \in R$ represents a random variable $x_i$ with a conditional probability table $p_i(x_i | x_{\text{pa}(i)})$. Each decision node $i \in D$ represents a controllable decision variable $x_i$, whose value is determined by a decision maker via a decision rule (or policy) $\delta_i: \mathbb{X}_{\text{pa}(i)} \to \mathbb{X}_i$, which determines the values of $x_i$ based on the observation on the values of $x_{\text{pa}(i)}$; we call the collection of policies $\boldsymbol{\delta} = \{\delta_i | i \in D\}$ a *strategy*. Finally, a utility function $u: \mathbb{X} \to \mathbb{R}^+$ measures the reward given an instantiation of $\boldsymbol{x} = [x_R, x_D]$, which the decision maker wants to maximize. It is reasonable to assume some decomposition structure on the utility $u(\boldsymbol{x})$, either additive, $u(\boldsymbol{x}) = \sum_{j \in U} u_j(x_{\beta_j})$, or multiplicative, $u(\boldsymbol{x}) = \prod_{j \in U} u_j(x_{\beta_j})$. A decomposable utility function can be visualized by augmenting the DAG with a set of leaf nodes $U$, called *utility nodes*, each with parent set $\beta_j$. See Fig. 1 for a simple example.

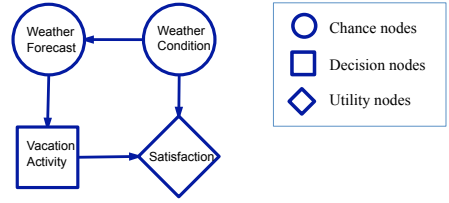

Figure 1: A simple influence diagram for deciding vacation activity [Shachter, 2007].

A decision rule $\delta_i$ is alternatively represented as a deterministic conditional "probability" $p_i^\delta(x_i | x_{\text{pa}(i)})$, where $p_i^\delta(x_i | x_{\text{pa}(i)}) = 1$ for $x_i = \delta_i(x_{\text{pa}(i)})$ and zero otherwise. It is helpful to allow *soft decision rules* where $p_i^\delta(x_i | x_{\text{pa}(i)})$ takes fractional values; these define a *randomized strategy* in which $x_i$ is determined by randomly drawing from $p_i^\delta(x_i | x_{\text{pa}(i)})$. We denote by $\Delta^o$ the set of deterministic strategies and $\Delta$ the set of randomized strategies. Note that $\Delta^o$ is a discrete set, while $\Delta$ is its convex hull.

Given an influence diagram, the optimal strategy should maximize the *expected utility function* (MEU):

$$\text{MEU} = \max_{\boldsymbol{\delta} \in \Delta} \text{EU}(\boldsymbol{\delta}) = \max_{\boldsymbol{\delta} \in \Delta} \mathbb{E}(u(\boldsymbol{x}) | \boldsymbol{\delta})$$
$$= \max_{\boldsymbol{\delta} \in \Delta} \sum_{\boldsymbol{x}} u(\boldsymbol{x}) \prod_{i \in C} p_i(x_i | x_{\text{pa}(i)}) \prod_{i \in D} p_i^\delta(x_i | x_{\text{pa}(i)})$$
$$\stackrel{def}{=} \max_{\boldsymbol{\delta} \in \Delta} \sum_{\boldsymbol{x}} \exp(\theta(\boldsymbol{x})) \prod_{i \in D} p_i^\delta(x_i | x_{\text{pa}(i)}) \quad (4)$$

where $\theta(\boldsymbol{x}) = \log[u(\boldsymbol{x}) \prod_{i \in C} p_i(x_i | x_{\text{pa}(i)})]$; we call the distribution $q(\boldsymbol{x}) \propto \exp(\theta(\boldsymbol{x}))$ the *augmented distribution* [Bielza et al., 1999]. The concept of the augmented distribution is critical since it completely specifies a MEU problem without the semantics of the influence diagram; hence one can specify $q(\boldsymbol{x})$ arbitrarily, e.g., via an undirected MRF, extending the definition of IDs. We can treat MEU as a special sort of "inference" on the augmented distribution, which as we will show, generalizes more common inference tasks.

In (4) we maximize the expected utility over $\Delta$; this is equivalent to maximizing over $\Delta^o$, since

**Lemma 2.1.** *For any ID*, $\max_{\boldsymbol{\delta} \in \Delta} EU(\boldsymbol{\delta}) = \max_{\boldsymbol{\delta} \in \Delta^o} EU(\boldsymbol{\delta})$.

**Perfect Recall Assumption.** The MEU problem can be solved in closed form if the influence diagram satisfies a *perfect recall assumption* (PRA) — there exists a "temporal" ordering over all the decision nodes, say $\{d_1, d_2, \cdots, d_m\}$, consistent with the partial order defined by the DAG $G$, such that every decision node observes all the earlier decision nodes and their parents, that is, $\{d_j\} \cup \text{pa}(d_j) \subseteq \text{pa}(d_i)$ for any $j < i$. Intuitively, PRA implies a centralized decision scenario, where a global decision maker sets all the decision nodes in a predefined order, with perfect memory of all the past observations and decisions.

With PRA, the chance nodes can be grouped by when they are observed. Let $r_{i-1}$ ($i = 1, \ldots, m$) be the set of chance nodes that are parents of $d_i$ but not of any $d_j$ for $j < i$; then both decision and chance nodes are ordered by $o = \{r_0, d_1, r_1, \cdots, d_m, r_m\}$. The MEU and its optimal strategy for IDs with PRA can be calculated by a sequential sum-max-sum rule,

$$\text{MEU} = \sum_{x_{r_0}} \max_{x_{d_1}} \sum_{x_{r_1}} \cdots \max_{x_{d_m}} \sum_{x_{r_m}} \exp(\theta(\boldsymbol{x})), \quad (5)$$

$$\delta^*_{d_i}(x_{\text{pa}(d_i)}) = \arg\max_{x_{d_i}} \Big\{ \sum_{x_{r_i}} \cdots \max_{x_{d_m}} \sum_{x_{r_m}} \exp(\theta(\boldsymbol{x})) \Big\},$$

where the calculation is performed in reverse temporal ordering, interleaving marginalizing chance nodes and maximizing decision nodes. Eq. (5) generalizes the inference tasks in Section 2.1, arbitrarily interleaving the sum and max operators. For example, marginal MAP can be treated as a *blind decision problem*, where no chance nodes are observed by any decision nodes.

As in other inference tasks, the calculation of the sum-max-sum rule can be organized into local computations if the augmented distribution $q(\boldsymbol{x})$ is factorized. However, since the max and sum operators are not exchangeable, the calculation of (5) is restricted to elimination orders consistent with the "temporal ordering". Notoriously, this "constrained" tree-width can be very high even for trees. See Koller and Friedman [2009].

However, PRA is often unrealistic. First, most systems lack enough memory to express arbitrary policies over an entire history of observations. Second, many practical scenarios, like team decision analysis [Detwarasiti and Shachter, 2005] and decentralized sensor networks [Kreidl and Willsky, 2006], are distributed by nature: a team of agents makes decisions independently based on sharing limited information with their neighbors. In these cases, relaxing PRA is very important.

**Imperfect Recall.** General IDs with the perfect recall assumption relaxed are discussed in Zhang et al. [1994], Lauritzen and Nilsson [2001], Koller and Milch [2003], and are commonly referred as limited memory influence diagrams (LIMIDs). Unfortunately, the re-laxation causes many difficulties. First, it is no longer possible to eliminate the decision nodes in a sequential "sum-max-sum" fashion. Instead, the dependencies of the decision nodes have cycles, formally discussed in Koller and Milch [2003] by defining a relevance graph over the decision nodes; the relevance graph is a tree with PRA, but is usually loopy with imperfect recall. Thus iterative algorithms are usually required for LIMIDs. Second and more importantly, the incomplete information may cause the agents to behave myopically, selfishly choosing locally optimal strategies due to ignorance of the global statistics. This breaks the strategy space into many local modes, making the MEU problem non-convex; see Fig. 2 for an illustration.

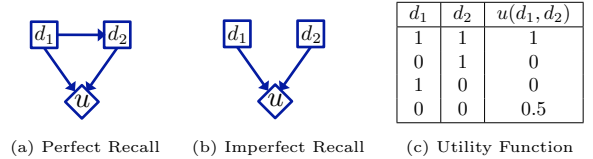

(a) Perfect Recall  (b) Imperfect Recall  (c) Utility Function

Figure 2: Illustrating imperfect recall. In (a) $d_2$ observes $d_1$; its optimal decision rule is to equal $d_1$'s state (whatever it is); knowing $d_2$ will follow, $d_1$ can choose $d_1 = 1$ to achieve the global optimum. In (b) $d_1$ and $d_2$ do not know the other's states; both $d_1 = d_2 = 1$ and $d_1 = d_2 = 0$ (suboptimal) become locally optimal strategies and the problem is multi-modal.

The most popular algorithms for LIMIDs are based on policy-by-policy improvement, e.g., the *single policy update* (SPU) algorithm [Lauritzen and Nilsson, 2001] sequentially optimizes $\delta_i$ with $\boldsymbol{\delta}_{\neg i} = \{\delta_j : j \neq i\}$ fixed:

$$\delta_i(x_{\text{pa}(i)}) \leftarrow \arg\max_{x_i} \mathbb{E}(u(\boldsymbol{x})|x_{\text{fam}(i)}; \boldsymbol{\delta}_{\neg i}), \quad (6)$$

$$\mathbb{E}(u(\boldsymbol{x})|x_{\text{fam}(i)}; \boldsymbol{\delta}_{\neg i}) = \sum_{x_{\neg\text{fam}(i)}} \exp(\theta(\boldsymbol{x})) \prod_{j \in D \setminus \{i\}} p^\delta_j(x_j | x_{\text{pa}(j)}),$$

where $\text{fam}(i) = \{i\} \cup \text{pa}(i)$. The update circles through all $i \in D$ in some order, and ties are broken arbitrarily in case of multiple maxima. The expected utility in SPU is non-decreasing at each iteration, and it gives a locally optimal strategy at convergence in the sense that the expected utility can not be improved by changing any single node's policy. Unfortunately, SPU's solution is heavily influenced by initialization and can be very suboptimal.

This issue is helped by generalizing SPU to the strategy improvement (SI) algorithm [Detwarasiti and Shachter, 2005], which simultaneously updates subgroups of decisions nodes. However, the time and space complexity of SI grows exponentially with the sizes of subgroups. In the sequel, we present a novel variational framework for MEU, and propose BP-like algorithms that go beyond the naïve greedy paradigm.

## 3 Duality Form of MEU

In this section, we derive a duality form for MEU, generalizing the duality results of the standard inference in Section 2.1. Our main result is summarized in the following theorem.

**Theorem 3.1.** *(a). For an influence diagram with augmented distribution $q(\bm{x}) \propto \exp(\theta(\bm{x}))$, its log maximum expected utility $\log \text{MEU}(\bm{\theta})$ equals*

$$\max_{\bm{\tau} \in \mathbb{M}} \{\langle \bm{\theta}, \bm{\tau} \rangle + H(\bm{x}; \bm{\tau}) - \sum_{i \in D} H(x_i | x_{\text{pa}(i)}; \bm{\tau})\}. \quad (7)$$

*Suppose $\bm{\tau}^*$ is a maximum of (7), then $\bm{\delta}^* = \{\bm{\tau}^*(x_i | x_{\text{pa}(i)}) | i \in D\}$ is an optimal strategy.*

*(b). For IDs with perfect recall, (7) reduces to*

$$\max_{\bm{\tau} \in \mathbb{M}} \{\langle \bm{\theta}, \bm{\tau} \rangle + \sum_{o_i \in C} H(x_{o_i} | x_{o_{1:i-1}}; \bm{\tau})\}, \quad (8)$$

*where $o$ is the temporal ordering of the perfect recall.*

*Proof.* (a) See appendix; (b) note PRA implies $\text{pa}(i) = o_{1:i-1}$ ($i \in D$), and apply the entropy chain rule. □

The distinction between (8) and (7) is subtle but important: although (8) (with perfect recall) is always (if not strictly) a convex optimization, (7) (without perfect recall) may be non-convex if the subtracted entropy terms overwhelm; this matches the intuition that incomplete information sharing gives rise to multiple locally optimal strategies.

The MEU duality (8) for ID with PRA generalizes earlier duality results of inference: with no decision nodes, $D = \emptyset$ and (8) reduces to (1) for the log-partition function; when $C = \emptyset$, no entropy terms appear and (8) reduces to the linear program relaxation of MAP. Also, (8) reduces to marginal MAP when no chance nodes are observed before any decision. As we show in Section 4, this unification suggests a line of unified algorithms for all these different inference tasks.

Several corollaries provide additional insights.

**Corollary 3.2.** *For an ID with parameter $\bm{\theta}$, we have*

$$\log \text{MEU} = \max_{\bm{\tau} \in \mathbb{I}} \{\langle \bm{\theta}, \bm{\tau} \rangle + \sum_{o_i \in C} H(x_{o_i} | x_{o_{1:i-1}}; \bm{\tau})\} \quad (9)$$

*where $\mathbb{I} = \{\bm{\tau} \in \mathbb{M} \colon x_{o_i} \perp x_{o_{1:i-1} \setminus \text{pa}(o_i)} | x_{\text{pa}(o_i)}, \forall o_i \in D\}$, corresponding to those distributions that respect the imperfect recall constraints; "$x \perp y | z$" denotes conditional independence of $x$ and $y$ given $z$.*

Corollary 3.2 gives another intuitive interpretation of imperfect recall vs. perfect recall: MEU with imperfect recall optimizes same objective function, but over a subset of the marginal polytope that restricts the observation domains of the decision rules; this non-convex inner subset is similar to the mean field approximation for partition functions. See Wolpert [2006] for a similar connection to mean field for bounded rational game theory. Interestingly, this shows that extending a LIMID to have perfect recall (by extending the observation domains of the decision nodes) can be considered a "convex" relaxation of the LIMID.

**Corollary 3.3.** *For any $\epsilon$, if $\bm{\tau}^*$ is global optimum of*

$$\max_{\bm{\tau} \in \mathbb{M}} \{\langle \bm{\theta}, \bm{\tau} \rangle + H(\bm{x}) - (1 - \epsilon) \sum_{i \in D} H(x_i | x_{\text{pa}(i)})\}. \quad (10)$$

*and $\bm{\delta}^* = \{\tau^*(x_i | x_{\text{pa}(i)}) | i \in D\}$ is a deterministic strategy, then it is an optimal strategy for MEU.*

The parameter $\epsilon$ is a temperature to "anneal" the MEU problem, and trades off convexity and optimality. For large $\epsilon$, e.g., $\epsilon \geq 1$, the objective in (10) is a strictly convex function, while $\bm{\delta}^*$ is unlikely to be deterministic nor optimal (if $\epsilon = 1$, (10) reduces to standard marginazation); as $\epsilon$ decreases towards zero, $\bm{\delta}^*$ becomes more deterministic, but (10) becomes more non-convex and is harder to solve. In Section 4 we derive several possible optimization approaches.

## 4 Algorithms

The duality results in Section 3 offer new perspectives for MEU, allowing us to bring the tools of variational inference to develop new efficient algorithms. In this section, we present a junction graph framework for BP-like MEU algorithms, and provide theoretical analysis. In addition, we propose two double-loop algorithms that alleviate the issue of non-convexity in LIMIDs or provide convergence guarantees: a deterministic annealing approach suggested by Corollary 3.3 and a method based on the proximal point algorithm.

### 4.1 A Belief Propagation Algorithm

We start by formulating the problem (7) into the junction graph framework. Let $(\mathcal{G}, \mathcal{C}, \mathcal{S})$ be a junction graph for the augmented distribution $q(\bm{x}) \propto \exp(\theta(\bm{x}))$. For each decision node $i \in D$, we associate it with exactly one cluster $c_k \in \mathcal{C}$ satisfying $\{i, \text{pa}(i)\} \subseteq c_k$; we call such a cluster a *decision cluster*. The clusters $\mathcal{C}$ are thus partitioned into decision clusters $\mathcal{D}$ and the other (normal) clusters $\mathcal{R}$. For simplicity, we assume each decision cluster $c_k \in \mathcal{D}$ is associated with exactly one decision node, denoted $d_k$.

Following the junction graph framework in Section 2.1, the MEU dual (10) (with temperature parameter $\epsilon$) is

approximated by

$$\max_{\boldsymbol{\tau}\in\mathbb{L}}\{\langle\boldsymbol{\theta},\boldsymbol{\tau}\rangle+\sum_{k\in\mathcal{R}}H_{c_k}+\sum_{k\in\mathcal{D}}H^\epsilon_{c_k}-\sum_{(kl)\in\mathcal{E}}H_{s_{kl}}\}, \quad (11)$$

where $H_{c_k} = H(x_{c_k})$, $H_{s_{kl}} = H(x_{s_{kl}})$ and $H^\epsilon(x_{c_k}) = H(x_{c_k}) - (1-\epsilon)H(x_{d_k}|x_{\text{pa}(d_k)})$. The dependence of entropies on $\boldsymbol{\tau}$ is suppressed for compactness. Eq. (11) is similar to the objective of regular sum-product junction graph BP, except the entropy terms of the decision clusters are replaced by $H^\epsilon_{c_k}$.

Using a Lagrange multiplier method similar to Yedidia et al. [2005], a hybrid message passing algorithm can be derived for solving (11):

**Sum messages:**
(normal clusters) $\quad m_{k\to l} \propto \sum_{x_{c_k\setminus s_{kl}}} \psi_{c_k} m_{\sim k\setminus l}, \quad (12)$

**MEU messages:**
(decision clusters) $\quad m_{k\to l} \propto \sum_{x_{c_k\setminus s_{kl}}} \frac{\sigma_k[\psi_{c_k} m_{\sim k}; \epsilon]}{m_{l\to k}}, \quad (13)$

where $\sigma_k[\cdot]$ is an operator that solves an annealed local MEU problem associated with $b(x_{c_k}) \propto \psi_{c_k} m_{\sim k}$:

$$\sigma_k[b(x_{c_k}); \epsilon] \stackrel{def}{=} b(x_{c_k}) b_\epsilon(x_{d_k}|x_{\text{pa}(d_k)})^{1-\epsilon}$$

where $b_\epsilon(x_{d_k}|x_{\text{pa}(d_k)})$ is the "annealed" optimal policy

$$b_\epsilon(x_{d_k}|x_{\text{pa}(d_k)}) = \frac{b(x_{d_k}, x_{\text{pa}(d_k)})^{1/\epsilon}}{\sum_{x_{d_k}} b(x_{d_k}, x_{\text{pa}(d_k)})^{1/\epsilon}},$$

$$b(x_{d_k}, x_{\text{pa}(d_k)}) = \sum_{x_{z_k}} b(x_{c_k}), \quad z_k = c_k \setminus \{d_k, \text{pa}(d_k)\}.$$

As $\epsilon \to 0^+$, one can show that $b_\epsilon(x_{d_k}|x_{\text{pa}(d_k)})$ is exactly an optimal strategy of the local MEU problem with augmented distribution $b(x_{c_k})$.

At convergence, the stationary point of (11) is:

$$\tau_{c_k} \propto \psi_{c_k} m_{\sim k} \quad \text{for normal clusters} \quad (14)$$
$$\tau_{c_k} \propto \sigma_k[\psi_{c_k} m_{\sim k}; \epsilon] \quad \text{for decision clusters} \quad (15)$$
$$\tau_{s_{kl}} \propto m_{k\to l} m_{l\to k} \quad \text{for separators} \quad (16)$$

This message passing algorithm reduces to sum-product BP when there are no decision clusters. The outgoing messages from decision clusters are the crucial ingredient, and correspond to solving local (annealed) MEU problems.

Taking $\epsilon \to 0^+$ in the MEU message update (13) gives a fixed point algorithm for solving the original objective directly. Alternatively, one can adopt a deterministic annealing approach [Rose, 1998] by gradually decreasing $\epsilon$, e.g., taking $\epsilon^t = 1/t$ at iteration $t$.

**Reparameterization Properties.** BP algorithms, including sum-product, max-product, and hybrid message passing, can often be interpreted as reparameterization operators, with fixed points satisfying some sum (resp. max or hybrid) consistency property yet leaving the joint distribution unchanged [e.g., Wainwright et al., 2003a, Weiss et al., 2007, Liu and Ihler, 2011]. We define a set of "MEU-beliefs" $\boldsymbol{b} = \{b(x_{c_k}), b(x_{s_{kl}})\}$ by $b(x_{c_k}) \propto \psi_{c_k} m_k$ for all $c_k \in \mathcal{C}$, and $b(x_{s_{kl}}) \propto m_{k\to l} m_{l\to k}$; note that the "beliefs" $\boldsymbol{b}$ are distinguished from the "marginals" $\boldsymbol{\tau}$. We can show that at each iteration of MEU-BP in (12)-(13), the $\boldsymbol{b}$ satisfy

**Reparameterization:** $\quad q(\boldsymbol{x}) \propto \frac{\prod_{k\in\mathcal{V}} b(x_{c_k})}{\prod_{(kl)\in\mathcal{E}} b(x_{s_{kl}})}, \quad (17)$

and further, at a fixed point of MEU-BP we have

**Sum-consistency:**
(normal clusters) $\quad \sum_{c_k\setminus s_{ij}} b(x_{c_k}) = b(x_{s_{kl}}), \quad (18)$

**MEU-consistency:**
(decision clusters) $\quad \sum_{c_k\setminus s_{ij}} \sigma_k[b(x_{c_k}); \epsilon] = b(x_{s_{kl}}). \quad (19)$

**Optimality Guarantees.** Optimality guarantees of MEU-BP (with $\epsilon \to 0^+$) can be derived via reparameterization. Our result is analogous to those of Weiss and Freeman [2001] for max-product BP and Liu and Ihler [2011] for marginal-MAP.

For a junction tree, a tree-order is a partial ordering on the nodes with $k \preceq l$ iff the unique path from a special cluster (called root) to $l$ passes through $k$; the parent $\pi(k)$ is the unique neighbor of $k$ on the path to the root. Given a subset of decision nodes $D'$, a junction tree is said to be *consistent* for $D'$ if there exists a tree-order with $s_{k,\pi(k)} \subseteq \text{pa}(d_k)$ for any $d_k \in D'$.

**Theorem 4.1.** *Let $(\mathcal{G}, \mathcal{C}, \mathcal{S})$ be a consistent junction tree for a subset of decision nodes $D'$, and $\boldsymbol{b}$ be a set of MEU-beliefs satisfying the reparameterization and consistency conditions (17)-(19) with $\epsilon \to 0^+$. Let $\boldsymbol{\delta}^* = \{b_{c_k}(x_{d_k}|x_{\text{pa}(d_k)}): d_k \in D\}$; then $\boldsymbol{\delta}^*$ is a locally optimal strategy in the sense that $\text{EU}(\{\boldsymbol{\delta}^*_{D'}, \boldsymbol{\delta}_{D\setminus D'}\}) \leq \text{EU}(\boldsymbol{\delta}^*)$ for any $\boldsymbol{\delta}_{D\setminus D'}$.*

A junction tree is said to be *globally* consistent if it is consistent for all the decision nodes, which as implied by Theorem 4.1, ensures a globally optimal strategy; this notation of *global consistency* is similar to the *strong junction trees* in Jensen et al. [1994]. For IDs with perfect recall, a globally consistent junction tree can be constructed by a standard procedure which triangulates the DAG of the ID along reverse temporal order. For IDs without perfect recall, it is usually not possible to construct a globally consistent junction tree; this is the case for the toy example in Fig. 2b. However, coordinate-wise optimality follows as a consequence of Theorem 4.1 for general IDs with arbitrary junction trees, indicating that MEU-BP is at least as "optimal" as SPU.

**Theorem 4.2.** *Let $(\mathcal{G}, \mathcal{C}, \mathcal{S})$ be an arbitrary junction tree, and $\boldsymbol{b}$ and $\boldsymbol{\delta}^*$ defined in Theorem 4.1. Then $\boldsymbol{\delta}^*$ is a locally person-by-person optimal strategy: $\mathrm{EU}(\{\boldsymbol{\delta}_i^*, \boldsymbol{\delta}_{D\setminus i}\}) \leq \mathrm{EU}(\boldsymbol{\delta}^*)$ for any $i \in D$ and $\boldsymbol{\delta}_{D\setminus i}$.*

**Additively Decomposable Utilities.** Our algorithms rely on the factorization structure of the augmented distribution $q(\boldsymbol{x})$. For this reason, multiplicative utilities fit naturally, but additive utilities are more difficult (as they also are in exact inference) [Koller and Friedman, 2009]. To create factorization structure in additive utility problems, we augment the model with a latent "selector" variable, similar to that in mixture models. For details, see the appendix.

### 4.2 Proximal Algorithms

In this section, we present a proximal point approach [e.g., Martinet, 1970, Rockafellar, 1976] for the MEU problems. Similar methods have been applied to standard inference problems, e.g., Ravikumar et al. [2010].

We start with a brief introduction to the proximal point algorithm. Consider an optimization problem $\min_{\boldsymbol{\tau} \in \mathbb{M}} f(\boldsymbol{\tau})$. A proximal method instead iteratively solves a sequence of "proximal" problems

$$\boldsymbol{\tau}^{t+1} = \arg\min_{\boldsymbol{\tau} \in \mathbb{M}} \{f(\boldsymbol{\tau}) + w^t D(\boldsymbol{\tau}||\boldsymbol{\tau}^t)\}, \qquad (20)$$

where $\boldsymbol{\tau}^t$ is the solution at iteration $t$ and $w^t$ is a positive coefficient. $D(\cdot||\cdot)$ is a distance, called the proximal function; typical choices are Euclidean or Bregman distances or $\psi$-divergences [e.g., Teboulle, 1992, Iusem and Teboulle, 1993]. Convergence of proximal algorithms has been well studied: the objective series $\{f(\boldsymbol{\tau}^t)\}$ is guaranteed to be non-increasing at each iteration, and $\{\boldsymbol{\tau}^t\}$ converges to an optimal solution (sometimes superlinearly) for convex programs, under some regularity conditions on the coefficients $\{w^t\}$. See, e.g., Rockafellar [1976], Tseng and Bertsekas [1993], Iusem and Teboulle [1993].

Here, we use an entropic proximal function that naturally fits the MEU problem:

$$D(\boldsymbol{\tau}||\boldsymbol{\tau}') = \sum_{i \in D} \sum_{\boldsymbol{x}} \tau(\boldsymbol{x}) \log[\tau_i(x_i|x_{\mathrm{pa}(i)})/\tau_i'(x_i|x_{\mathrm{pa}(i)})],$$

a sum of conditional KL-divergences. The proximal update for the MEU dual (7) then reduces to

$$\boldsymbol{\tau}^{t+1} = \arg\max_{\boldsymbol{\tau} \in \mathbb{M}} \{\langle \boldsymbol{\theta}^t, \boldsymbol{\tau} \rangle + H(\boldsymbol{x}) - (1-w^t)H(x_i|x_{\mathrm{pa}(i)})\}$$

where $\theta^t(\boldsymbol{x}) = \theta(\boldsymbol{x}) + w^t \sum_{i \in D} \log \tau_i^t(x_i|x_{\mathrm{pa}(i)})$. This has the same form as the annealed problem (10) and can be solved by the message passing scheme (12)-(13). Unlike annealing, the proximal algorithm updates $\boldsymbol{\theta}^t$ each iteration and does not need $w^t$ to approach zero.

We use two choices of coefficients $\{w^t\}$: (1) $w^t = 1$ (*constant*), and (2) $w^t = 1/t$ (*harmonic*). The choice $w^t = 1$ is especially interesting because the proximal update reduces to a standard *marginalization* problem, solvable by standard tools without the MEU's temporal elimination order restrictions. Concretely, the proximal update in this case reduces to

$$\tau_i^{t+1}(x_i|x_{\mathrm{pa}(i)}) \propto \tau_i^t(x_i|x_{\mathrm{pa}(i)}) \mathbb{E}(u(\boldsymbol{x})|x_{\mathrm{fam}(i)}; \boldsymbol{\delta}_{\neg i}^n)$$

with $\mathbb{E}(u(\boldsymbol{x})|x_{\mathrm{fam}(i)}; \boldsymbol{\delta}_{\neg i}^n)$ as defined in (6). This proximal update can be seen as a "soft" and "parallel" version of the greedy update (6), which makes a hard update at a single decision node, instead of a soft modification simultaneously for all decision nodes. The soft update makes it possible to correct earlier suboptimal choices and allows decision nodes to make cooperative movements. However, convergence with $w^t = 1$ may be slow; using $w^t = 1/t$ takes larger steps but is no longer a standard marginalization.

## 5 Experiments

We demonstrate our algorithms on several influence diagrams, including randomly generated IDs, large scale IDs constructed from problems in the UAI08 inference challenge, and finally practically motivated IDs for decentralized detection in wireless sensor networks. We find that our algorithms typically find better solutions than SPU with comparable time complexity; for large scale problems with many decision nodes, our algorithms are more computationally efficient than SPU because one step of SPU requires updating (6) (a global expectation) for all the decision nodes.

In all experiments, we test single policy updating (SPU), our MEU-BP running directly at zero temperature (BP-0$^+$), annealed BP with temperature $\epsilon^t = 1/t$ (Anneal-BP-1/t), and the proximal versions with $w^t = 1$ (Prox-BP-one) and $w^t = 1/t$ (Prox-BP-1/t). For the BP-based algorithms, we use two constructions of junction graphs: a standard junction tree by triangulating the DAG in backwards topological order, and a loopy junction graph following [Mateescu et al., 2010] that corresponds to Pearl's loopy BP; for SPU, we use the same junction graphs to calculate the inner update (6). The junction trees ensure the inner updates of SPU and Prox-BP-one are performed exactly, and has optimality guarantees in Theorem 4.1, but may be computationally more expensive than the loopy junction graphs. For the proximal versions, we set a maximum of 5 iterations in the inner loop; changing this value did not seem to lead to significantly different results. The BP-based algorithms may return non-deterministic strategies; we round to deterministic strategies by taking the largest values.

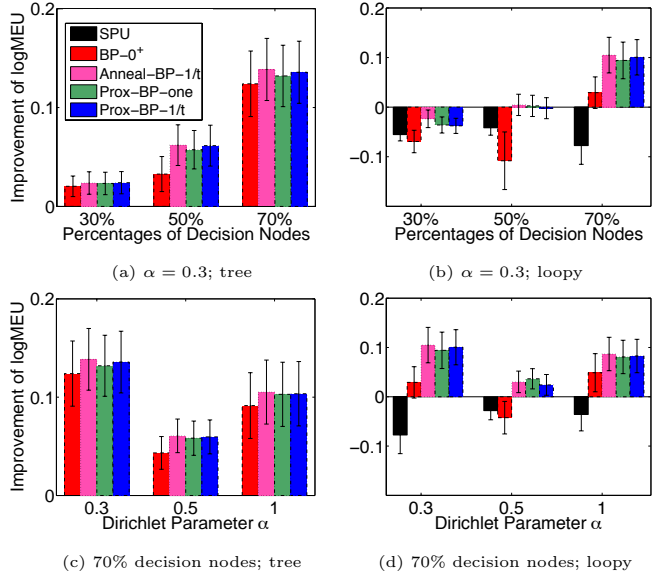

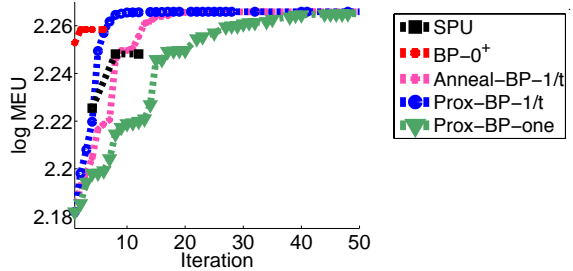

Figure 4: A typical trajectory of MEU (of the rounded deterministic strategies) v.s. iterations for the random IDs in Fig. 3. One iteration of the BP-like methods denotes a forward-backward reduction on the junction graph; One step of SPU requires $|D|$ (number of decision nodes) reductions. SPU and BP-$0^+$ are stuck at a local model in the 2nd iteration.

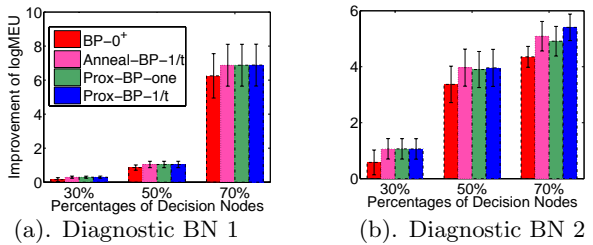

Figure 5: Results on IDs constructed from two diagnostic BNs from the UAI08 challenge. Here all algorithms used the loopy junction graph and are initialized uniformly. (a)-(b) the logMEU of algorithms normalized to that of SPU. Averaged on 10 trails.

Figure 3: Results on random IDs of size 20. The y-axes show the log MEU of each algorithm compared to SPU on a junction tree. The left panels correspond to running the algorithms on junction trees, and right panels on loopy junction graphs. (a) & (b) shows MEUs as the percentage of decision nodes changes. (c) & (d) show MEUs v.s. the Dirichlet parameter $\alpha$. The results are averaged on 20 random models.

**Random Bayesian Networks.** We test our algorithms on randomly constructed IDs with additive utilities. We first generate a set of random DAGs of size 20 with maximum parent size of 3. To create IDs, we take the leaf nodes to be utility nodes, and among non-leaf nodes we randomly select a fixed percentage to be decision nodes, with the others being chance nodes. We assume the chance and decision variables are discrete with 4 states. The conditional probability tables of the chance nodes are randomly drawn from a symmetric Dirichlet distribution $\mathrm{Dir}(\alpha)$, and the entries of the utility function from Gamma distribution $\Gamma(\alpha, 1)$.

The relative improvement of log MEU compared to the SPU with junction tree are reported in Fig. 3. We find that when using junction trees, all our BP-based methods dominate SPU; for loopy junction graphs, BP-$0^+$ occasionally performs worse than SPU, but all the annealed and proximal algorithms outperform SPU with the same loopy junction graph, and often even SPU with junction tree. As the percentage of decision nodes increases, the improvement of the BP-based methods on SPU generally increases. Fig. 4 shows a typical trajectory of the algorithms across iterations. The algorithms were initialized uniformly; random initializations behaved similarly, but are omitted for space.

**Diagnostic Bayesian networks.** We construct larger scale IDs based on two diagnostic Bayes nets with 200-300 nodes and 300-600 edges, taken from the UAI08 inference challenge. To create influence diagrams, we made the leaf nodes utilities, each defined by its conditional probability when clamped to a randomly chosen state, and total utility as the product of the local utilities (multiplicatively decomposable). The set of decision nodes is again randomly selected among the non-leaf nodes with a fixed percentage. Since the network sizes are large, we only run the algorithms on the loopy junction graphs. Again, our algorithms significantly improve on SPU; see Fig. 5.

**Decentralized Sensor Network.** In this section, we test an influence diagram constructed for decentralized detection in wireless sensor networks [e.g., Viswanathan and Varshney, 1997, Kreidl and Willsky, 2006]. The task is to detect the states of a hidden process $p(h)$ (as a pairwise MRF) using a set of distributed sensors; each sensor provides a noisy measurement $v_i$ of the local state $h_i$, and overall performance is boosted by allowing the sensor to transmit small (1-bit) signals $s_i$ along an directional path, to help the predictions of their downstream sensors. The utility function includes rewards for correct prediction and a cost for sending signals. We construct an ID as sketched in Fig. 6(a) for addressing the *offline* policy design task, finding optimal policies of how to predict the states based on the local measurement and received signals, and policies of whether and how to pass signals to downstream nodes; see appendix for more details.

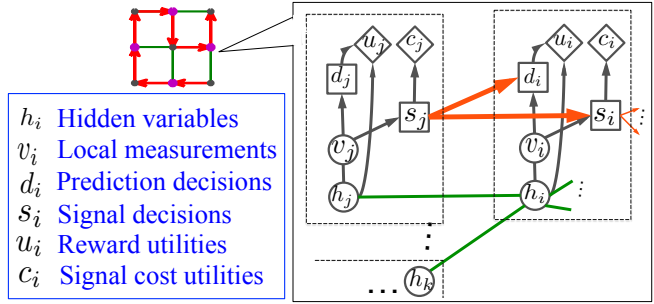
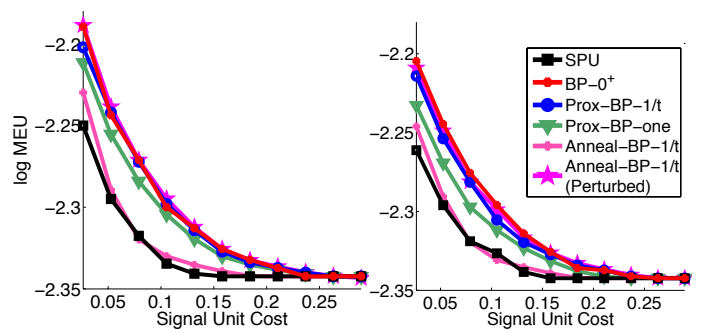

(a) The ID for sensor network detection  (b) Junction tree  (c) Loopy junction graph

Figure 6: (a) A sensor network on $3 \times 3$ grid; green lines denote the MRF edges of the hidden process $p(h)$, on some of which (red arrows) signals are allowed to pass; each sensor may be accurate (purple) or noisy (black). Optimal strategies should pass signals from accurate sensors to noisy ones but not the reverse. (b)-(c) The log MEU of algorithms running on (b) a junction tree and (c) a loopy junction graph. As the signal cost increases, all algorithms converge to the communication-free strategy. Results averaged on 10 random trials.

To escape the "all-zero" fixed point, we initialize the proximal algorithms and SPU with 5 random policies, and BP-$0^+$ and `Anneal-BP-1/t` with 5 random messages. We first test on a sensor network on a $3 \times 3$ grid, where the algorithms are run on both a junction tree constructed by standard triangulation and a loopy junction graph (see the Appendix for construction details). As shown in Fig. 6(b)-(c), SPU performs worst in all cases. Interestingly, `Anneal-BP-1/t` performs relatively poorly here, because the annealing steps make it insensitive to and unable to exploit the random initializations; this can be fixed by a "perturbed" annealed method that injects a random perturbation into the model, and gradually decreases the perturbation level across iterations (`Anneal-BP-1/t (Perturbed)`).

A similar experiment (with only the loopy junction graph) is performed on the larger random graph in Fig. 7; the algorithm performances follow similar trends. SPU performs even worse in this case since it appears to over-send signals when two "good" sensors connect to one "bad" sensor.

## 6 Related Works

Many exact algorithms for ID have been developed, usually in a variable-elimination or message-passing form; see Koller and Friedman [2009] for a recent review. Approximation algorithms are relatively unex-plored, and usually based on separately approximating individual components of exact algorithms [e.g., Sabbadin et al., 2011, Sallans, 2003]; our method instead builds an integrated framework. Other approaches, including MCMC [e.g., Charnes and Shenoy, 2004] and search methods [e.g., Marinescu, 2010], also exist but are usually more expensive than SPU or our BP-like methods. See the appendix for more discussion.

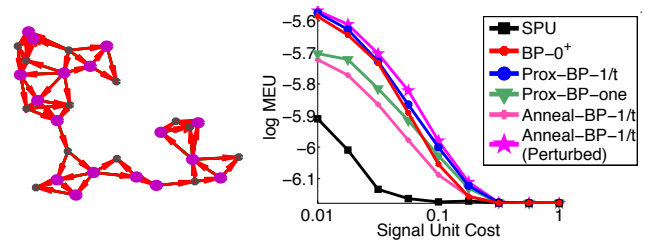

(a) Sensor network  (b) Loopy junction graph

Figure 7: The results on a sensor network on a random graph with 30 nodes (the MRF edges overlap with the signal paths). Averaged on 5 random models.

## 7 Conclusion

In this work we derive a general variational framework for influence diagrams, for both the "convex" centralized decisions with perfect recall and "non-convex" decentralized decisions. We derive several algorithms, but equally importantly open the door for many others that can be applied within our framework. Since these algorithms rely on *decomposing* the global problems into local ones, they also open the possibility of efficiently distributable algorithms.

**Acknowledgements.** Work supported in part by NSF IIS-1065618 and a Microsoft Research Fellowship.

# References

C. Bielza, P. Muller, and D. R. Insua. Decision analysis by augmented probability simulation. *Management Science*, 45(7):995–1007, 1999.

J. Charnes and P. Shenoy. Multistage Monte Carlo method for solving influence diagrams using local computation. *Management Science*, pages 405–418, 2004.

A. Detwarasiti and R. D. Shachter. Influence diagrams for team decision analysis. *Decision Anal.*, 2, Dec 2005.

R. Howard and J. Matheson. Influence diagrams. In *Readings on Principles & Appl. Decision Analysis*, 1985.

R. Howard and J. Matheson. Influence diagrams. *Decision Anal.*, 2(3):127–143, 2005.

A. Iusem and M. Teboulle. On the convergence rate of entropic proximal optimization methods. *Computational and Applied Mathematics*, 12:153–168, 1993.

F. Jensen, F. V. Jensen, and S. L. Dittmer. From influence diagrams to junction trees. In *UAI*, pages 367–373. Morgan Kaufmann, 1994.

J. Jiang, P. Rai, and H. Daumé III. Message-passing for approximate MAP inference with latent variables. In *NIPS*, 2011.

D. Koller and N. Friedman. *Probabilistic graphical models: principles and techniques*. MIT press, 2009.

D. Koller and B. Milch. Multi-agent influence diagrams for representing and solving games. *Games and Economic Behavior*, 45(1):181–221, 2003.

O. P. Kreidl and A. S. Willsky. An efficient message-passing algorithm for optimizing decentralized detection networks. In *IEEE Conf. Decision Control*, Dec 2006.

S. Lauritzen and D. Nilsson. Representing and solving decision problems with limited information. *Managment Sci.*, pages 1235–1251, 2001.

Q. Liu and A. Ihler. Variational algorithms for marginal MAP. In *UAI*, Barcelona, Spain, July 2011.

R. Marinescu. A new approach to influence diagrams evaluation. In *Research and Development in Intelligent Systems XXVI*, pages 107–120. Springer London, 2010.

B. Martinet. Régularisation d'inéquations variationnelles par approximations successives. *Revue Française dInformatique et de Recherche Opérationelle*, 4:154–158, 1970.

R. Mateescu, K. Kask, V. Gogate, and R. Dechter. Join-graph propagation algorithms. *Journal of Artificial Intelligence Research*, 37:279–328, 2010.

J. Pearl. Influence diagrams - historical and personal perspectives. *Decision Anal.*, 2(4):232–234, dec 2005.

P. Ravikumar, A. Agarwal, and M. J. Wainwright. Message-passing for graph-structured linear programs: Proximal projections, convergence, and rounding schemes. *Journal of Machine Learning Research*, 11:1043–1080, Mar 2010.

R. T. Rockafellar. Monotone operators and the proximal point algorithm. *SIAM Journal on Control and Optimization*, 14(5):877, 1976.

K. Rose. Deterministic annealing for clustering, compression, classification, regression, and related optimization problems. *Proc. IEEE*, 86(11):2210 –2239, Nov 1998.

R. Sabbadin, N. Peyrard, and N. Forsell. A framework and a mean-field algorithm for the local control of spatial processes. *International Journal of Approximate Reasoning*, 2011.

B. Sallans. Variational action selection for influence diagrams. Technical Report OEFAI-TR-2003-29, Austrian Research Institute for Artificial Intelligence, 2003.

R. Shachter. Model building with belief networks and influence diagrams. *Advances in decision analysis: from foundations to applications*, page 177, 2007.

M. Teboulle. Entropic proximal mappings with applications to nonlinear programming. *Mathematics of Operations Research*, 17(3):pp. 670–690, 1992.

P. Tseng and D. Bertsekas. On the convergence of the exponential multiplier method for convex programming. *Mathematical Programming*, 60(1):1–19, 1993.

R. Viswanathan and P. Varshney. Distributed detection with multiple sensors: part I – fundamentals. *Proc. IEEE*, 85(1):54 –63, Jan 1997.

M. Wainwright and M. Jordan. Graphical models, exponential families, and variational inference. *Found. Trends Mach. Learn.*, 1(1-2):1–305, 2008.

M. Wainwright, T. Jaakkola, and A. Willsky. A new class of upper bounds on the log partition function. *IEEE Trans. Info. Theory*, 51(7):2313–2335, July 2005.

M. J. Wainwright, T. Jaakkola, and A. S. Willsky. Tree-based reparameterization framework for analysis of sum-product and related algorithms. *IEEE Trans. Info. Theory*, 45:1120–1146, 2003a.

M. J. Wainwright, T. S. Jaakkola, and A. S. Willsky. MAP estimation via agreement on (hyper) trees: Message-passing and linear programming approaches. *IEEE Trans. Info. Theory*, 51(11):3697 – 3717, Nov 2003b.

Y. Weiss and W. Freeman. On the optimality of solutions of the max-product belief-propagation algorithm in arbitrary graphs. *IEEE Trans. Info. Theory*, 47(2):736 –744, Feb 2001.

Y. Weiss, C. Yanover, and T. Meltzer. MAP estimation, linear programming and belief propagation with convex free energies. In *UAI*, 2007.

D. Wolpert. Information theory – the bridge connecting bounded rational game theory and statistical physics. *Complex Engineered Systems*, pages 262–290, 2006.

J. Yedidia, W. Freeman, and Y. Weiss. Constructing free-energy approximations and generalized BP algorithms. *IEEE Trans. Info. Theory*, 51, July 2005.

N. L. Zhang. Probabilistic inference in influence diagrams. In *Computational Intelligence*, pages 514–522, 1998.

N. L. Zhang, R. Qi, and D. Poole. A computational theory of decision networks. *Int. J. Approx. Reason.*, 11:83–158, 1994.